\documentclass{article}

\usepackage[preprint]{neurips_2023}

\usepackage{amsmath,amsfonts,bm}

\def\eqref#1{equation~\ref{#1}}

\def\1{\bm{1}}

\DeclareMathAlphabet{\mathsfit}{\encodingdefault}{\sfdefault}{m}{sl}
\SetMathAlphabet{\mathsfit}{bold}{\encodingdefault}{\sfdefault}{bx}{n}

\newcommand{\E}{\mathbb{E}}

\DeclareMathOperator*{\argmax}{arg\,max}

\usepackage{hyperref}       
\usepackage{url}            
\usepackage{booktabs}       
\usepackage{nicefrac}       
\usepackage{xcolor}         
\usepackage{graphicx}
\usepackage{subcaption}
\usepackage{amsmath}
\usepackage{wrapfig}
\usepackage{cleveref}
\usepackage{makecell}
\usepackage{siunitx}
\usepackage{pifont}
\usepackage{setspace}
\usepackage{tabulary}
\usepackage{babel,adjustbox,booktabs,multirow}
\usepackage{cleveref}
\usepackage{threeparttable}
\usepackage{xspace}
\newcommand{\cmark}{\ding{51}}%
\newcommand{\xmark}{\ding{55}}%
\newcommand{\methodName}{ChemRLformer\xspace}%

\title{Searching for High-Value Molecules Using Reinforcement Learning and Transformers}

\author{
\begin{minipage}{\textwidth}
\centering
Raj Ghugare$^{1,2}$  \qquad Santiago Miret $^3$ \qquad Adriana Hugessen$^{1,2}$ \\
\vspace{0.3em}
\textbf{Mariano Phielipp$^3$ \qquad Glen Berseth$^{1,2}$} \\
\vspace{0.5em}
\normalfont $^1$Université de Montréal\qquad $^2$Mila - Quebec AI Institute \qquad $^3$Intel Labs  \\
\end{minipage}
}

\begin{document}

\maketitle

\begin{abstract}
  Reinforcement learning (RL) over text representations can be effective for finding high-value policies that can search over graphs. However, RL requires careful structuring of the search space and algorithm design to be effective in this challenge. 
  Through extensive experiments, we explore how different design choices for text grammar and algorithmic choices for training can affect an RL policy's ability to generate molecules with desired properties. We arrive at a new RL-based molecular design algorithm (\methodName) and perform a thorough analysis using 25 molecule design tasks, including computationally complex protein docking simulations. From this analysis, we discover unique insights in this problem space and show that \methodName achieves state-of-the-art performance while being more straightforward than prior work by demystifying which design choices are actually helpful for text-based molecule design.
\end{abstract}

\section{Introduction}
\label{introduction}

Molecular discovery can have a significant impact on our society, however, the vast search space makes it challenging to find high-value molecules.
The potential of reinforcement learning (RL) methods to discover new, high-value molecules has resulted in a series of research work performed by RL researchers focusing on learning policies as graph neural networks (GNNs)~\citep{you2018gcpn, zhou2019moldqn, jin2020multiobjective, fu2021moler, yang2021hit, bengio2021gfn}. In this formulation, the RL policy is trained to add atoms and bonds to a molecular graph representation.
In this formulation there is a one-to-one mapping between molecules and their graph representation, making it easier to construct state and action spaces with Markovian dynamics. However, the action space in the graph formulation is vast as it consists of the product of candidate attachment positions and candidate attachment sequences. 
Graph-based data structures (such as adjacency matrices, trees, etc.) are a powerful representation used to describe a number of design problems, including social networks \citep{tan2019deep}, transportation networks \citep{wang2021deep}, recommendation systems \citep{chen2021survey}, and combinatorial optimization problems \citep{khadka2020optimizing, miret2022neuroevolution} have been popular in this design space. However, GNNs are often difficult to train \citep{chen2022bag} and cannot readily take advantage of large-scale text data sets that effectively describe molecular structures and properties.

In order to take advantage of the richness of text-based representations for molecules, one can formulate the molecular search problem as the construction of tokens in a sequence that become a molecular text.
The molecular texts formulated by common text-based representations, such as SMILES \citep{weininger1988smiles} and SELFIES \citep{krenn2020selfies}, can then be converted into molecular graphs with cheminformatics libraries \citep{landrum2013rdkit} using their respective encoding and decoding rules. However, the text-based representation can be more difficult to formulate as an MDP since there is not always an exact one-to-one mapping between texts and molecules. In fact, the text-to-molecule conversion can be many-to-one, where the complexity of the dynamics in the MDP given by many-to-one mappings is non-trivial.
On the other hand, the action space in molecular text design can be significantly reduced given the rules of text construction imposed by a given representation. 
Moreover, formulating molecule discovery as sequence-generation has the potential to capitalize on recent successes in natural language modeling \citep{NEURIPS2020_1457c0d6}.

In this paper, we perform a detailed empirical study of molecular discovery using 
text-based RL across more than 25 molecular properties relevant for drug-discovery, including docking simulation for molecular ligands \citep{garcia2022dockstring, huang2022artificial} and develop our own algorithm (MoLRL) based on state-of-the-art literature as shown in \Cref{tab:rl-methods}. In our experiments, we evaluate two molecular text representations (SMILES, SELFIES) and the use of three neural network architectures (Multi-Layer Perceptron \citep{10.5555/944919.944966}, Recurrent Neural Network \citep{schmidt2019recurrent}, Transformer \citep{vaswani2017attention}) pretrained on 5 datasets of varying quality and sizes. We create \methodName that achieves the highest performance across these tasks while being much simpler than previous text-based RL algorithms~\citep{blaschke2020reinvent, gao2022pmo}.

Via our detailed ablation study, we construct \methodName and find that pretraining on \textit{aligned} datasets can significantly improve performance across all molecular design tasks, even exceeding the performance of agents pretrained on 100~times larger datasets. We also show that targeted algorithmic design, such as hill-climbing in the replay buffer and regularization, further increases the performance of \methodName. To the best of our knowledge, \methodName is the largest analysis of text-based RL methods for molecule discovery.  

\begin{table*}[ht]
\begin{spacing}{1.05}
	\centering
	\caption{Table showing conceptual comparisons of various text based molecular optimization methods. MoLRL combines the most successful elements of prior work.} 
	\label{tab:rl-methods}
	\vspace{0.5mm}
	\begin{adjustbox}{max width=1\linewidth}
        \begin{threeparttable}
	\begin{tabular}{cccccc}
        \toprule
		\bf{Method} & \bf{Text Representation} & \bf{RL} & \bf{Architecture} & \bf{Pretraining} & \bf{Algorithmic Components} \\

\midrule
\multicolumn{6}{c}{Literature Methods} \\
\midrule
SMILES-VAE~\citep{gomez2018automatic} & SMILES    & \xmark          & VAE  & \cmark & Maximum Likelihood \\
SMILES-LSTM~\citep{brown2019gmole} & SMILES    & \xmark          & LSTM  & \cmark &  Maximum Likelihood \\
BOSS~\citep{moss2020boss} & SMILES    & \xmark          & VAE     & \xmark  &  Bayesian Optimization \\
REINVENT~\citep{blaschke2020reinvent}  & SMILES    & \cmark   & GRU  & \cmark & Replay buffer, KL \\
REINVENT 2.0~\citep{reinvent_2.0} & SMILES    & \cmark       & GRU   & \cmark  & HC-Replay buffer, Log p, KL\\
STONED~\citep{nigam2021janus} & SELFIES    & \xmark          & FC  & \xmark  &  Genetic algorithm \\ 
Pasithea~\citep{Shen_2021} & SELFIES    & \xmark          & FC  & \xmark  &  Deep dreaming \\
\midrule
\textbf{\methodName (Ours)}  & SMILES, SELFIES    & \cmark   &  Transformer, FC  & \cmark & Replay buffer, KL \\

        \bottomrule
	\end{tabular}
        \end{threeparttable}
	\end{adjustbox}
\end{spacing}
\vspace{-1mm}
\end{table*}

\section{Related Work}

\textbf{RL for Design and Discovery}:
Many methods in diverse fields leverage RL to help augment a prior design method to improve performance~\citep{Yu2018-aq,Schaff2019-xy}. Other methods have explicitly included the design process in the RL loop by training design problems together~\citep{Chen2021-gu,Ha2019-zb,Luck2020-ff,kumardata} with  most prior work focusing on robot and agent design, not molecular design. Our molecular design work creates an autoregressive structure that grows the size of the state as the agent acts in the environment.

\textbf{Molecular Discovery Using Sequence-Based Methods:} Sequence-based methods treat molecular design as a sequence of tokens that get concatenated in order. Generative models for sequence-based methods span a diverse range, including variational autoencoders (VAEs) \citep{gomez2018automatic, alperstein2019all}, recurrent neural networks (RNNs) \citep{gupta2018generative, bjerrum2017molecular, grisoni2020bidirectional, flam2022language} and transformer models\citep{wang2019smiles, fabian2020molecular, edwards-etal-2022-translation, zeng2022deep, taylor2022galactica}. The general procedure for all the above methods is to perform self-supervised generative learning to sample molecules similar to the original dataset. MoLRL can also make use of pretrained generative models, which we then fine-tune using reinforcement learning to produce enhanced molecules.

\textbf{Molecular Discovery Using Search-Based Methods:} Although sequence-based molecule generation methods often provide a more structured way of learning molecular distributions, search-based methods generally have the advantage of being able to directly find molecules based on a desired property.
Although a wide range of graph-based RL methods~\citep{you2018gcpn, zhou2019moldqn, jin2020multiobjective, fu2021moler, yang2021hit, bengio2021gfn} for optimizing molecules exist, graph-based state representations introduce significant complexity to the RL problem formulation, both in the transition dynamics and action space. By contrast, text-based methods are simpler and also relatively under-explored, motivating our focus on these methods in this work. Moreover, recent work \citep{cieplinski2021able, gao2022pmo} has shown that an older text-based method REINVENT~\citep{olivecrona2017molecular} outperforms more complex graph-based RL methods. Some limited extensions to \citet{olivecrona2017molecular} have been explored, including experimenting with a newer molecular grammar designed for robust molecule generation \citep{gao2022pmo}. However, there has been limited work proposing the use of language models and text-based RL for molecular discovery. Additionally, there have been limited efforts to incorporate recent advancements from the language modeling domain into these methods. For example, the a character-level LSTM network architecture used in \citet{olivecrona2017molecular}, has not been revisited despite significant recent advances in sequence modeling \citep{vaswani2017attention, brown2020language}.

\section{Background}
\label{sec:background}

The algorithms detailed in this paper are built on top of a foundation of reinforcement learning, text-based molecule representations, and language modeling.

\paragraph{Reinforcement Learning:} 

Reinforcement learning can be used to learn policies for sequential decision-making problems. Policies are optimized based on an environment that is described as a Markov Decision Process (MDP). A discrete MDP is defined by the tuple $\langle\mathcal{S}, \mathcal{A}, \mathcal{T}, r, \gamma\rangle$ where $\mathcal{S}$ is the state space, $\mathcal{A}$ is the action space, $T: \mathcal{S} \times \mathcal{A} \times \mathcal{S'} \rightarrow [0,1]$ is the transition function, $r : \mathcal{S} \times \mathcal{A} \rightarrow \mathcal{R}$ is the reward function and $\gamma$ is the discount rate.

For actions $a_t \in \mathcal{A}$ and states $s_t \in \mathcal{S}$, the goal of reinforcement learning is to learn a policy $\pi_{\theta}(a_t|s_t)$ which maps states to actions, such that:
\begin{equation}
 	\pi_{\theta}(a_t|s_t) = \argmax_{\theta}\mathbf{E}_{p(\tau|\theta)}\left[\sum_{t=0}^T\gamma^tr(s_t,a_t)\right]
  	\label{eq:rl_obj}
 \end{equation}
where $p(\tau|\theta)$ is the distribution over trajectories induced by $\pi_{\theta}$ and the transition function $\mathcal{T}$.

\paragraph{Text representations for molecules:} 

Molecules are most naturally described using a graph structure of atoms and bonds. However, graph-based deep learning models can be difficult to train, especially at large scale~\citep{dwivedi2022benchmarking, geisler2023robustness}.
Recent works have proposed a variety of text representations for molecules~\citep{weininger1988smiles, krenn2020selfies, heller2013inchi, krenn2022selfies, cheng2023group}, each having their distinct advantages and shortcomings. In this study, we focus on the two most commonly used representations: SMILES~\citep{weininger1988smiles} and SELFIES~\citep{krenn2020selfies}. Any text representation for molecules consists of a set of valid tokens, which may represent individual atoms or special characters that imply the presence of certain structures,
as well as the encoding and decoding rules needed to convert between the text representation and the graph representation of a molecule. Valid texts under a grammar are those which respect both the vocabulary and the encoding/decoding rules for that grammar and, hence, can be converted into a graph representation of a molecule. SELFIES, which was developed in response to the tendency for SMILES-based deep learning models~\citep{G_mez_Bombarelli_2018, jin2018jtvae} to generate invalid molecular texts, has the useful property of providing a conversion for \textit{any} text into a graph corresponding to a molecule, provided the tokens in the text respect the SELFIES vocabulary. For example, the text representation of Benzene in SMILES is C1=CC=CC=C1 while in SELFIES one possible representation is [C][=C][C][=C][C][=C][Ring1][=Branch1].

\paragraph{Language modeling:} 
Language modeling often relies on the self-supervised task of next-token prediction for model pretraining. The general framework for next-token prediction is to train a model to predict the next token in a sequence autoregressively, i.e. given the previous tokens in the sequence (left context). Many architectures to handle sequential data have been proposed:
Recurrent Neural Networks (RNNs) \citep{schmidhuber1997lstm, Rumelhart1986LearningIR} are a class of models used in sequence modeling which use recursive connections in hidden layers to accumulate the left context for next-token prediction. Transformers are a more recent architecture that instead use a self-attention mechanism \citep{vaswani2017attention} to capture dependencies between all tokens in a sequence. For next-token prediction tasks, attention masking is used to enforce left context, meaning that representations for tokens later in the sequence are only allowed to attend to previous tokens in the sequence. In \Cref{sec:method} we outline how we pretrain an autoregressive sequence model to predict sequences of known molecules.

\section{\methodName Generating molecular strings via reinforcement learning}
\label{sec:method}

The molecular design space is complex but the benefit from finding improved options is great. In this section, we describe \methodName and how combining language models and tools from RL produces a sota algorithm.

\paragraph{MDP for molecule generation:} The vocabulary and grammar for text representations of molecules can be interpreted as an MDP as described in \Cref{sec:background}, where the states $s_t$ correspond to a \textit{variable length} text of accumulating tokens, and the actions $a_t$ correspond to vocabulary defined by the text-representation. The transition function is a deterministic function where the action $a_t$ taken by the agent is appended to the end of the state $s_t$ resulting in $s_{t+1}$ using the dynamics $ s_{t+1} = [s_t, a_t] \leftarrow T(s_t, a_t)$. 
However, the corresponding transition function induced in the graph representation of molecules is more complex as it is determined by the encoding/decoding rules of the chosen text representation. 
For example, in the SMILES grammar, a random concatenation of tokens may not correspond to a valid molecule, 
while the SELFIES grammar is constructed such that any ordering of its tokens is encoded as a valid molecule.

Finally, the reward function $\mathcal{R}$ scores molecules according to their alignment with desired chemical properties, which can involve complex material simulations. The underlying property computation of the reward function further informs the dynamics of the MDP imposed by text representation. For example, docking scores are used to estimate the binding affinity between ligands and protein targets. We discuss reward functions for molecules in more detail in section \Cref{sec:experiments}.

\begin{figure}[t]
    \centering
    \includegraphics[width=1\textwidth]{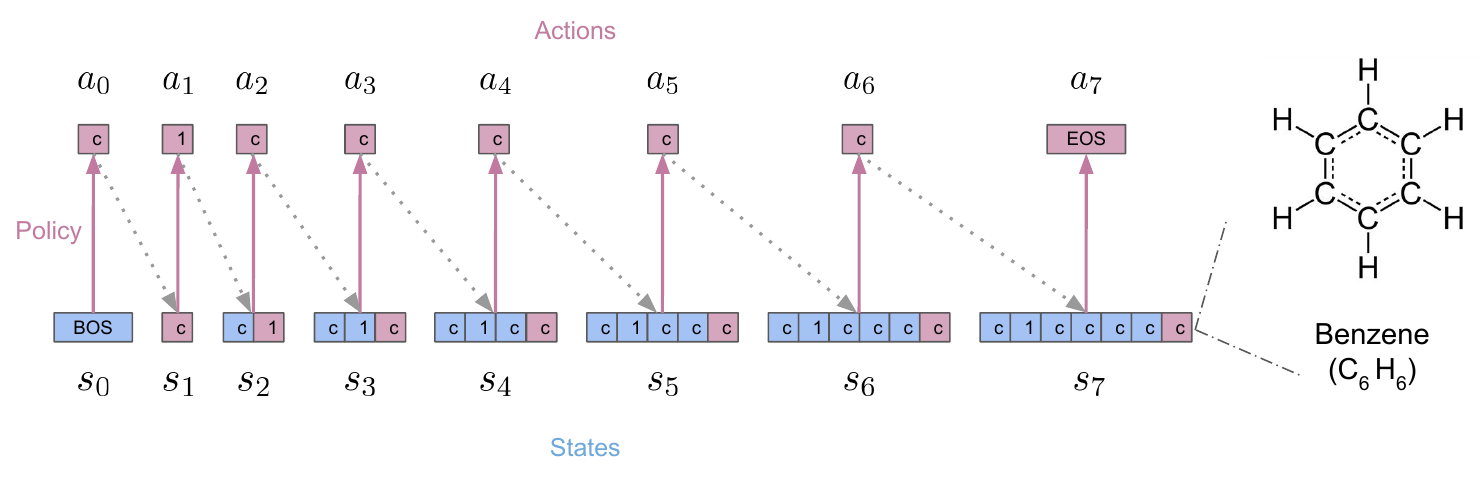}
    \caption{ \textbf{Autoregressively generating a benzene molecule.}: An autoregressive model for sequence generation can be viewed as an RL policy where the actions $a_t$ are the next tokens to append to the sequence and the state is the concatenation of all actions taken up to time $t-1$. A special end-of-sequence token can terminate the episode early at time $T$. The text at time $s_T$ is then converted into a molecule based on the text-representation grammar and then scored according to a scoring function that measures the alignment of the molecule with the desired properties informed by the application. Hydrogen atoms are added at the end to complete the structure.}
    \label{fig:ar-benzene-sampling}
\end{figure}

\paragraph{Pretraining policies for molecule discovery.}
To advance effectively within this vast search space, we make use of datasets containing a large number of drug-like molecules in text format~\citep{irwin2012zinc, sterling2015zinc, mendez2019chembl}. This data is used to train an autoregressive model to predict tokens that conform to the grammar for drug-like molecules, instead of the random texts that are generated from a randomly initialized policy, thereby significantly simplifying the exploration problem. In particular, we pretrain a network $p_{\phi}$ on the self-supervised objective of next-token prediction. Although large language models can be trained with other objectives, such as corrupted text reconstruction \citep{edwards2022translation}, these models are not a good fit for our purposes since they cannot generate diverse and valid molecules without access to carefully designed prompts.
{\footnotesize \begin{equation}
\label{eq:pretraining_obj}
  \min_{\theta}\E_{A \sim D} \left[ \sum_{t=1}^{H} - \log p_{\theta}(a_t=A_t \mid A_{t-1}, \cdots A_0) \right] .
\end{equation}}
In practice a minibatch of sequences $\{A^1, \cdots, A^m\}$ are sampled from the prior dataset~$D$ to evaluate the loss function in Equation \ref{eq:pretraining_obj}, and the parameters are trained using gradient descent. 

\subsection{RL for molecule generation}
To generate a molecule, a \methodName policy $\pi_{\theta}(a_t \mid s_t)$ is allowed to autoregressively sample tokens for a fixed number of timesteps $H$. The start state $s_0$ is always a beginning-of-sequence token $[\text{BOS}]$, and the agent can terminate early by taking the end-of-sequence action $[\text{EOS}]$. Figure~\ref{fig:ar-benzene-sampling}, shows how an RL policy can construct a Benzene molecule. Since we are only interested in the properties of the final molecule, there are no intermediate rewards and the goal of the RL policy is to maximize the expected scalar reward corresponding to the final constructed molecule, $r(s_T)$. 
Thus, assuming a discount rate $\gamma=1$, Equation \ref{eq:rl_obj} can be rewritten more simply as:
{\footnotesize \begin{align}
\label{eq:rl_mg_obj}
  & \max_{\theta}\E_{s_T \sim \pi_{\theta}}\left[ r(s_T) \right] \\
  & \text{where} \: s_T = [\text{BOS}][a_0] [a_1] \cdots [\text{EOS}], \: \text{is sampled autoregressively from the policy}\nonumber.
\end{align}}
Our experiments use the policy gradient algorithm~\citep{sutton1999reinforce} to train the RL policy because it is known to achieve state-of-the-art performance amongst RL for molecular optimization~\citep{olivecrona2017molecular}. 
Deep RL policies are able to learn the non-linear global structures of molecular texts which, as we show in section \ref{sec:experiments}, enables them to generalize to novel and diverse molecules. 
However, training RL policies from scratch is time-consuming and can make the exploration problem infeasibly difficult. Next, we explain how we adapt recent language modelling techniques to pretrain the RL policy. 

\paragraph{RL fine-tuning} The pretrained model can directly be used to sample novel drug-like molecules. These molecules, however, are not optimized for any particular property. Note that given our definition of the state $s_t$ as the concatenated history of all previous actions, this pretrained network is exactly analogous to the policy network in Equation \ref{eq:rl_obj}. Hence, by initializing $\pi_{\theta} = p_{\phi}$, and $\theta=\phi$, we can fine-tune this pretrained network by optimizing \autoref{eq:rl_obj} via the policy gradient algorithm - REINFORCE~\citep{NIPS1999_464d828b}. We need only to define a reward function $r(s_{T})$ which scores molecules according to their alignment with the desired properties.  
In the following experiments, we show that this fine-tuning is vital for \methodName to sample better molecules. We also highlight the importance of pretraining and study how the size and quality of the prior data affect the downstream ability of RL to search for high-value molecules.

\section{Experimental Results}
\label{sec:experiments}

Our proposed algorithm \methodName uses the best combinations of choices resulting from assessing the performance across three dimensions: (1) what pretraining factors are important to improve RL for molecular discovery (\Cref{result:pretraing-data}), (2) how the use of recent text-based molecule grammars facilitates downstream RL exploration (\Cref{result:string-representations-architectures}); and, lastly, (3) which specific algorithmic changes are necessary to improve RL performance (\Cref{result:algorithm-tricks}).

\subsection{Experimental Setup}
\label{sec:experiment-setup}

\paragraph{Tasks.} 
We evaluate \methodName against five different \textbf{docking} targets~\citep{amr2015qvina} (fa7, parp1, 5ht1b, jak2, and braf) previously explored in the literature~\citep{yang2021hit, lee2023exploring}. The docking scores used to estimate the binding affinity between ligands and protein targets are a complex function of the global molecular structure and have been proposed as chemically relevant benchmarks for molecule design algorithms~\citep{cieplinski2021able, tripp2022an}. In addition to the docking targets, we also evaluate on $22$ pharmaceutically-relevant oracle functions~\citep{huang2021therapeutics, gao2022pmo, brown2019gmole} (\textbf{pytdc} tasks), which include tasks such as optimizing proxies of bioactivity, similarity to target molecules, and combinations of multiple physiochemical drug properties. 

\paragraph{Evaluation metrics.}
We design our evaluation procedure with the final goal of identifying the best candidates to test in a wet lab. To discover such high-value candidate molecules, we use sota simulators that assign rewards to molecules by performing complex docking simulations~\citep{amr2015qvina} or using proxy models and chemical rules~\citep{huang2021therapeutics}.  
Previous works limit the number of molecules sampled during evaluation to around $3000$ for \textbf{docking} tasks~\citep{garcia2022dockstring, yang2021hit, lee2023exploring} and $10000$ for \textbf{pytdc} tasks~\citep{gao2022pmo, brown2019gmole} due to the computational cost associated with these reward simulators. We allow up to $25000$ unique oracle calls and up to $40000$ total oracle calls (allowing repeats).
We argue this better reflects the lower cost and availability of computing resources relative to wet-lab resources. From all the sampled molecules, the average score of the top-$k$  $(k=1,10,100)$ molecules is used as a performance metric. These top groups are an estimate of the algorithm's ability to discover a group of top-quality candidates that could be given to a wet lab for thorough testing. We report \textbf{pytdc} scores on a normalized basis between zero and one by default. Next, we normalize all docking scores by dividing them by -20 in our experiments. Additionally, we report \textit{diversity}, defined as the averaged internal distance of the top $100$ molecules, and \textit{redundancy}, defined as the total number of oracle calls that an agent makes for an already evaluated molecule.

\paragraph{Pretraining.}
We study how the quality and size of prior data affect the downstream RL performance of \methodName by pretraining a GPT~\citep{radford2018improving} style transformer model on five datasets of varying sizes and quality and using the pretrained model as an initialization for the RL agent's policy network.
See \Cref{table:pretraing-data} for the name, size, and description of all datasets used in our work. We also rank all datasets based on their quality on \textbf{docking} and \textbf{pytdc} tasks. We determine the quality of a dataset by the performance of molecules sampled from the model pretrained on that dataset.

The quality of \methodName's pretrained model is evaluated using the \textit{top-100} molecules sampled by the pretrained model under the same evaluation setup in \Cref{appendix:evaluation-metrics}. By default, these open-sourced datasets contain a large number of drug-like molecules in SMILES format. For our experiments, we also convert all datasets to the SELFIES format. Lastly, three different architectures are compared: \textbf{fully-connected (FC)}, \textbf{recurrent (RNN)} - a GRU and \textbf{transformer} - GPT style autoregressive model, and compare them on downstream RL tasks.
\begin{table*}[ht]
\begin{spacing}{1.05}
	\centering
	\caption{\textbf{Description of molecular datasets used for pretraining:} Datasets are ranked according to procedure described in \Cref{sec:experiment-setup}. Two datasets have the same rank if their average performance lies inside one standard error of the other. The datasets are drawn from a subset of the Zinc \citep{sterling2015zinc, irwin2022chemformer} and ChemBL \citep{Gaulton2012chembl} databases.}
	\label{table:pretraing-data}
	\vspace{0.5mm}
	\begin{adjustbox}{max width=1\linewidth}
        \begin{threeparttable}
	\begin{tabular}{l|c|c|c|c}
        \toprule
		\bf{Dataset} & \bf{Size} & \bf{Docking Rank} & \bf{Pytdc Rank} & \bf{Description} \\
\midrule
\makecell{
CHEMBL} & 1.2 M & 1 & 1 & \makecell{Manually curated database of bioactive\\molecules with drug-like properties~\citep{Gaulton2012chembl}.}\\
\midrule
\makecell{
ZINC 250K} & 250 K & 2 & 2 & \makecell{ ZINC database molecules curated for their\\pharmaceutical relevance and popularity~\citep{gao2022pmo}.} \\ 
\midrule
ZINC 1M & 1 M & 3 & 3  & Random molecules from $\approx$ 1.5 billion\\
ZINC 10M & 10 M & 3 & 4 &  molecules from the ZINC database~\citep{Sterling2015zinc15}.\\
\makecell{ ZINC 100M \\ } & 100 M & 3 & 4 & ZINC 1M $\subset$ ZINC 10M $\subset$ ZINC100M.\\
        \bottomrule
	\end{tabular}
        \end{threeparttable}
	\end{adjustbox}
\end{spacing}
\vspace{-1em}
\end{table*}

All of our experiments on \textbf{pytdc} tasks are run across 5 random seeds.
Since docking simulations are expensive and time consuming, we run all \textbf{docking} experiments across 3 random seeds. Experiments with different seeds use the same pretrained model which is only pretrained once for every dataset. Additional details about the task rewards, evaluation metrics, and the pretraining datasets and models are discussed in Appendix~\ref{appendix:tasks},~\ref{appendix:evaluation-metrics}, and ~\ref{appendix:pretraining} respectively.

\subsection{How does prior data affect the final performance of \methodName?}
\label{result:pretraing-data}

\begin{figure}[ht]
\vspace*{-0em}                                        
    \centering
    \begin{subfigure}[b]{1\textwidth}
    \centering
    \includegraphics[trim={0.0cm 0.0cm 0.0cm 0.0cm},clip,width=0.9\textwidth]{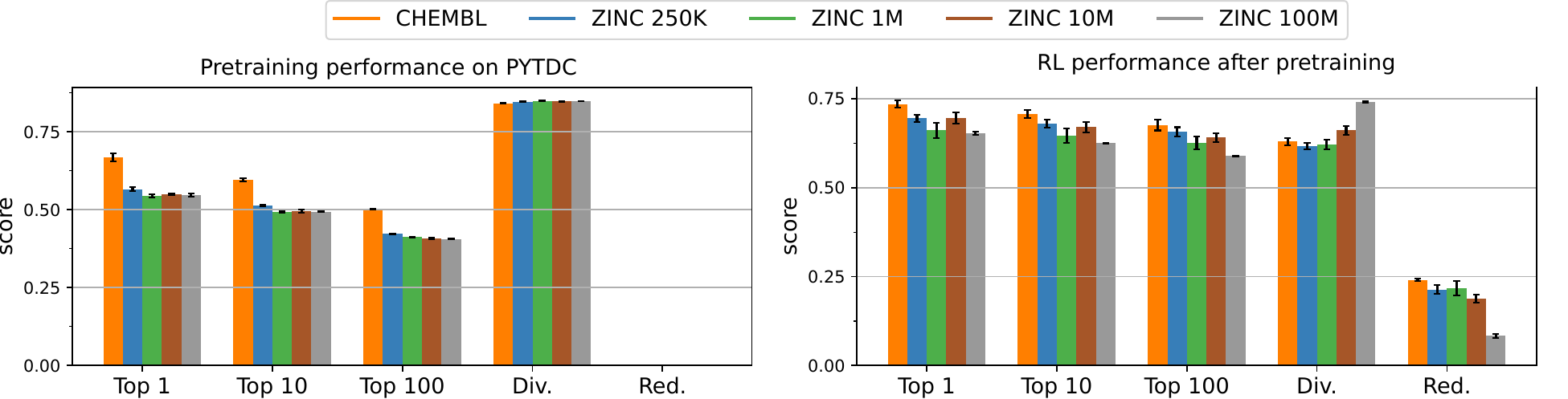}
    \subcaption{\label{fig:prior_data_rl_smiles} Performance on SMILES-based molecular design with pertaining (left) and with pretrianing and RL (right).}
    \end{subfigure}
    \begin{subfigure}[b]{1\textwidth}
    \centering
    \includegraphics[trim={0.0cm 0.0cm 0.0cm 1.0cm},clip,width=0.9\textwidth]{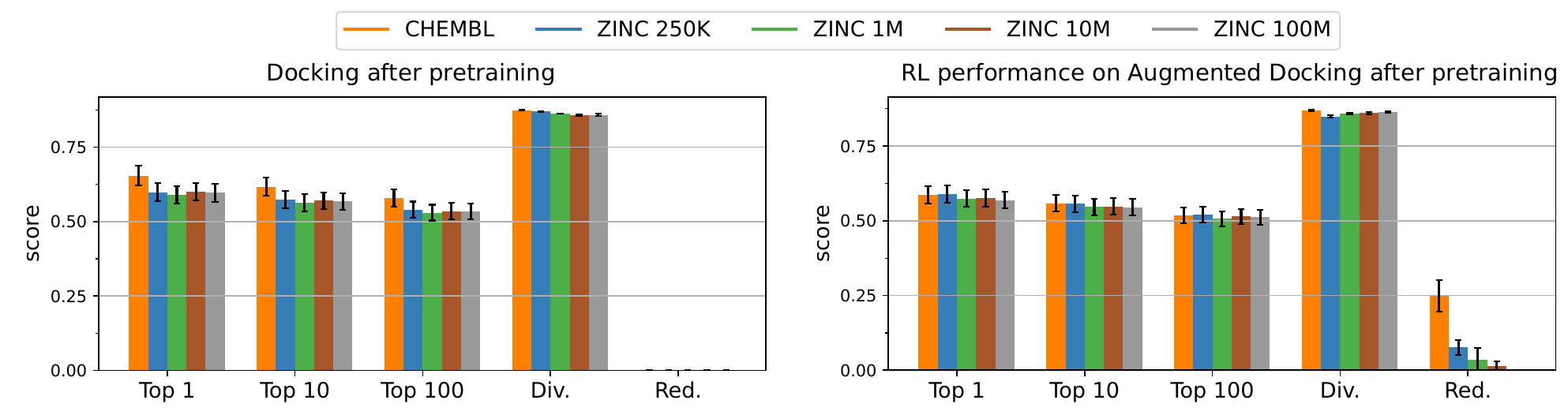}
    \subcaption{\label{fig:prior_data_rl_smiles} performance on SMILES-based molecular docking with pertaining (right) and with pretrainig and RL (right). \Cref{result:pretraing-data} describes augmented docking setting with additional experiments shown in~\Cref{appendix:pretraining}.}
    \end{subfigure}
   
    \caption{\label{fig:prior-data} 
    \label{fig:prior_data_pretrained_smiles} On the left pretrained performance on SMILES-based \methodName.
    Higher-quality datasets, such as ChemBL lead to higher-performance for both \textbf{pytdc} and augmented docking. On the right is the performance after RL training. RL has a substantial benefit for \textbf{pytdc} tasks, while for docking tasks an \textit{augmented docking} score is used to avoid reward hacking, see \Cref{fig:architecture_smiles} for details.}
    \vspace{-1em}
\end{figure}

In this section, we pretrain the REINFORCE policy on datasets of varying size and quality from~\Cref{table:pretraing-data}. Our datasets vary from small (250K) to very large (100M) sizes. Due to the parallelizability of training on larger datasets, we use the transformer policy architecture for all experiments in this section. In natural language processing (NLP), pretraining transformer models on large and diverse unlabelled datasets have been found to perform well on downstream tasks using few-shot labeled data~\citep{brown2020language}. Yet, our results in ~\Cref{fig:prior_data_rl_smiles} indicate that the quality of the prior dataset matters more than its size. \Cref{fig:prior_data_pretrained_smiles}, shows that the distribution of the \textit{ChEMBL} dataset is more aligned with both the \textbf{pytdc} and the \textbf{docking} tasks. As a result, the RL agent pretrained on the \textit{ChEMBL} dataset outperforms all other agents, including the ones trained on 100 times more data. 

Results may seem surprising from an NLP perspective, but they make sense when viewed from an RL perspective. Pretraining using next token prediction~\Cref{eq:pretraining_obj}, is analogous to behavior cloning in this context, where the performance depends largely on the quality of the offline dataset~\citep{ross2011reduction, ho2016generative}. These results suggest that \methodName might benefit from better pretraining objectives, that go beyond simple imitation learning, when trained on large and diverse offline datasets~\citep{kumar2023offline, farebrother2023protovalue}.

\subsection{Text representations and architectures for \methodName}
\label{result:string-representations-architectures}
Starting with a REINFORCE agent, we isolate the effect of various text representations for molecules and policy network architectures on performance. 
All experiments in this section use \textbf{ZINC-250k} dataset for pretraining. Similar results obtained for other datasets are shared in the following sections. Whenever we show normalized results across different experiments, we add the individual plots in \Cref{appendix:string-representations-architectures}. 

\begin{figure}[ht]
    \centering

    \includegraphics[width=0.9\textwidth]{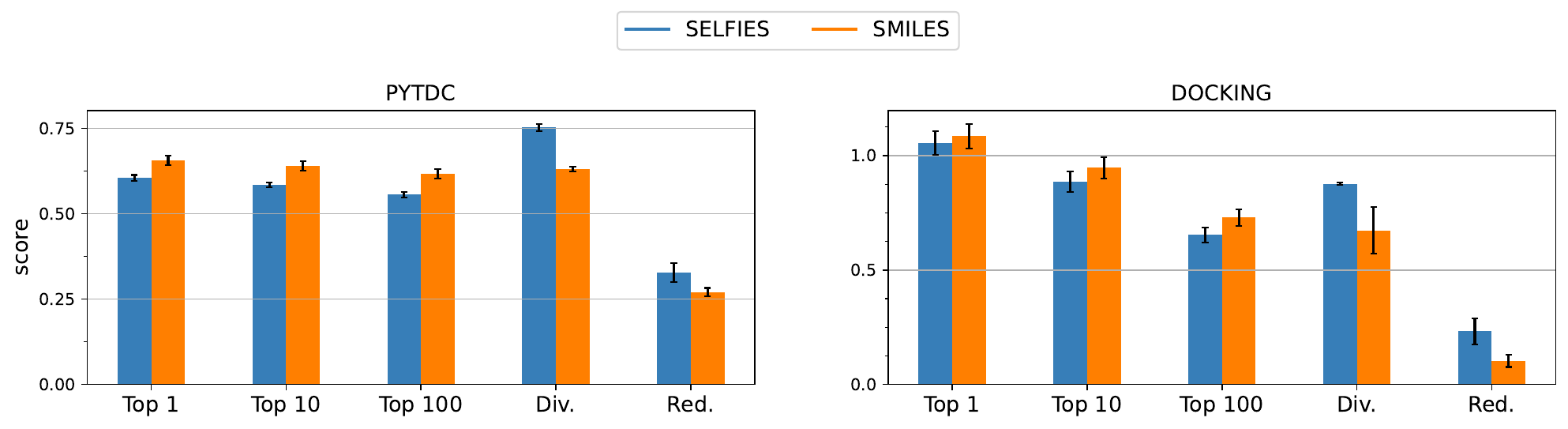}
    \caption{\label{fig:smiles-vs-selfies} 
    \textbf{Comparison between SELFIES and SMILES:} The SELFIES representation makes it relatively difficult for \methodName agents to explore effectively leading to generally lower performance on pytdc and docking while scoring higher on diversity. Scores are reported for the transformer model and are averaged across all reward functions.}
\end{figure}

\paragraph{Text representations.} In \Cref{fig:smiles-vs-selfies} we compare \methodName agents using different architectures and tasks across environments that base their dynamics on SELFIES and SMILES. The results show normalized scores across all architectures. Consistent with prior work \citep{gao2022pmo} we find that SMILES-based polices generally outperforms SELFIES-based policies. On all \textbf{pytdc} tasks and architectures, \methodName agents based on SMILES consistently achieve better rewards when compared to SELFIES-based agents across all reward metrics.
Although more subtle, we observe a similar theme in the \textbf{docking} tasks where SMILES achieves higher rewards than SELFIES on all top-K metrics. Another consistent theme in the results is that even though the diversity of top-100 molecules obtained by SELFIES is higher, the redundancy of SELFIES agents is higher as well. This means that SELFIES-based \methodName agents explore a much smaller region of the molecular space. These results suggest that the rules which allow SELFIES strings to always be converted into a valid molecule can actually be detrimental to the agent's exploration and search for high-value molecules, more details in \Cref{appendix:string-representations-architectures}. 

\begin{figure}[htb]
\vspace{-1em}
    \centering
    \includegraphics[width=.85\textwidth]{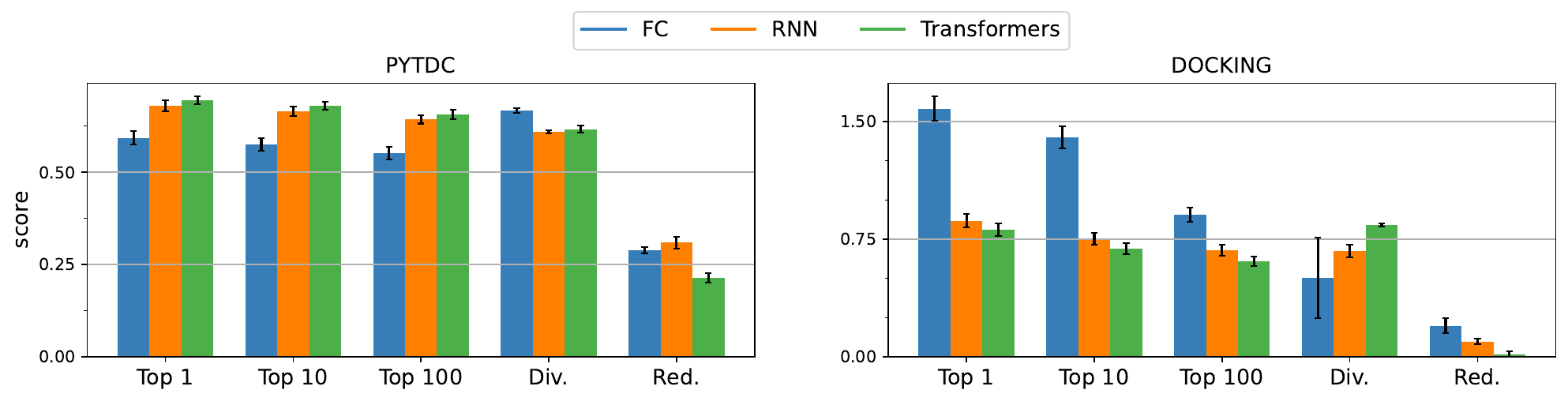}
    \caption{\label{fig:architecture_smiles} \textbf{Comparison of different policy architectures:} No single architecture clearly outperforms for molecular \methodName. Although FC does better on the docking tasks, our analysis shows that it learns to exploit the docking function as opposed to designing high-value molecules. More details about ways to tackle this issue are given in \Cref{appendix:reward-hacking}. Additional experiments for comparing transformers and RNNs are shown in \Cref{appendix:transformer-instability-rl}. These experiments use the smiles text representation.}
    \vspace{-1em}
\end{figure}

\paragraph{Architectures.} The results in \Cref{fig:architecture_smiles} show that the \textbf{transformer} and \textbf{RNN} have similar performance on all tasks. On the \textbf{pytdc} tasks, \textbf{FC} achieves worse performance than other architectures specially made to handle strings, as expected. However, on \textbf{docking} tasks, \textbf{FC} obtains unusually high rewards. We find that this method performs a type of \textit{reward function hacking}~\citep{amodei2016concrete,skalse2022defining,everitt2019towards} by exploiting a corner case of the docking-based reward function which provides high rewards for long strings of Carbon and Nitrogen atoms together. To evade the reward hacking of docking scores, we constructed an \textit{augmented docking} score function with commonly used oracles (QED and SA scores) based on previous work~\citep{lee2023exploring} (See~\Cref{appendix:pretraining} for more details).
This finding shows that the REINFORCE agent can search the space well and, in this case, can be used to expose issues with the current design of reward functions.

\subsection{Revisiting RL Algorithm Design Choices for \methodName.}
\label{result:algorithm-tricks}

Previous experiments identify key design and algorithmic choices for efficient text-based RL. But recent text-based RL algorithms~\citep{olivecrona2017molecular, bjerrum2023faster, thomas2022ahc} also employ many additional algorithmic components like replay buffers, hill climbing, KL regularisation towards the pretrained policy, and likelihood penalties. We perform an ablation study across these components to understand which ones are beneficial for the performance of \methodName. All experiments in this section are performed on \textbf{pytdc} tasks, using an RNN architecture as it is most commonly used in text-based RL. See~\Cref{appendix:algo-comp} for experiments with other architectures and reward functions.

\begin{wrapfigure}[10]{R}{0.5\textwidth}
    \centering
    \vspace{-2em}
    \includegraphics[width=0.5\textwidth]{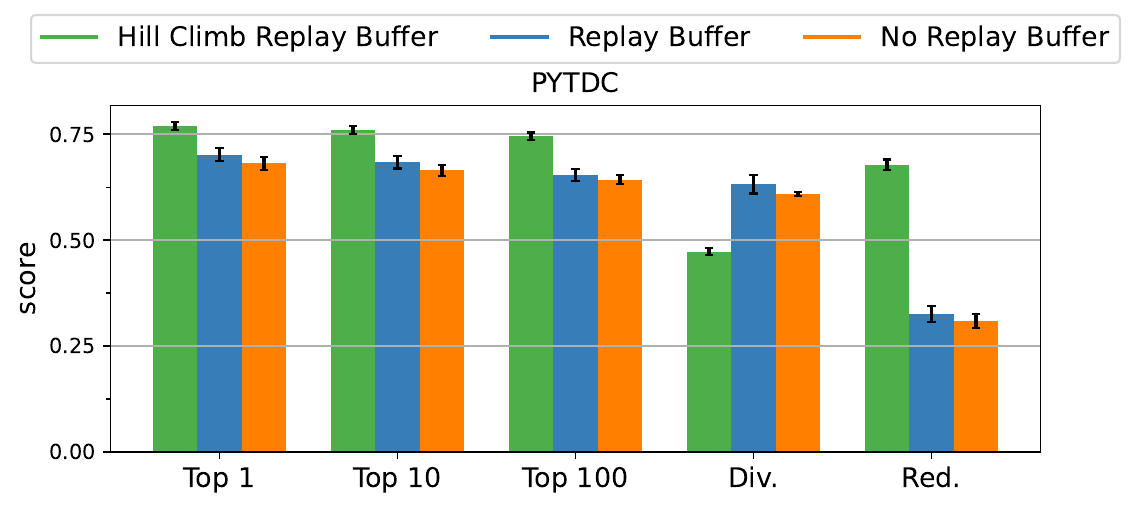}
    \vspace{-2em}
    \caption{\small Hill climbing buffer lead to 13\% improvement in Top-100 rewards.}
    \label{fig:replay-buffer}
\end{wrapfigure} 
\paragraph{Replay buffers and hill climbing.} In off-policy deep RL, a replay buffer is generally used to store and reuse previous trajectories for training. Although text-based RL algorithms are trained on-policy, prior work has proposed using a replay buffer to improve performance~\citep{mnih2013playing}. Standard replay buffers throw away the oldest trajectories as newer ones arrive. But many text-based RL algorithms propose to use hill-climb replay buffers, that randomly sample a batch of molecules from the highest scoring molecules seen so far and add them to the current mini batch. In \Cref{fig:replay-buffer}, we see that using the hill-climb buffer results in a significant performance boost for \methodName, whereas using a standard buffer does not contribute much. Notably, the use of a hill-climb replay buffer reduces diversity and increases redundancy quite substantially. The following two experiments involve combining regularisation terms with the RL objective in Equation ~\ref{eq:rl_mg_obj}. The coefficients on these extra terms can largely affect the final performance. To make a fair comparison, we perform hyper-parameter tuning over six different values for every new regularisation term with details provided in~\Cref{appendix:hp-tuning}.

\begin{wrapfigure}[10]{R}{0.5\textwidth}
    \centering
    \vspace{-2em}
    \includegraphics[width=\linewidth]{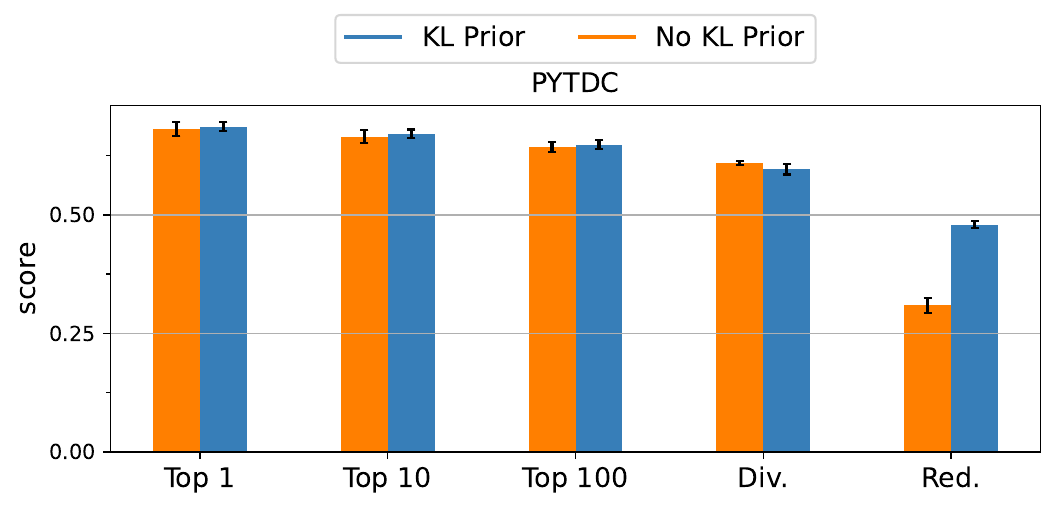}
    \vspace{-2em}
    \caption{\small Our experiments show little difference in performance for multiple KL regularization terms.}
    \label{fig:prior-kl}
\end{wrapfigure}
\paragraph{Should the policy to stay close to the pretrained model?} Pretrained models carry information on how to build valid drug-like molecules. To ensure that \methodName agents do not stray far away from the space of valid drug-like molecules during exploration, \citet{olivecrona2017molecular, gao2022pmo} constrain the KL divergence between the policy and the pretrained model by adding a KL penalty to the policy gradient loss function in ~\Cref{eq:rl_mg_obj}. Prior works show that adding this penalty helps the agent achieve better sample efficiency~\citep{gao2022pmo}. Yet, our results in~\autoref{fig:prior-kl} suggest that, when you increase the number of oracle calls in simulation, adding this penalty does not yield any additional benefit while substantially increasing the GPU memory requirement, especially when using larger models. Since invalid molecules correspond to zero rewards, the \methodName agent is able to learn to avoid invalid structures on its own merit. 

\begin{wrapfigure}{R}{0.4\textwidth}
    \centering
    \vspace{-2em}
    \includegraphics[width=1\linewidth]{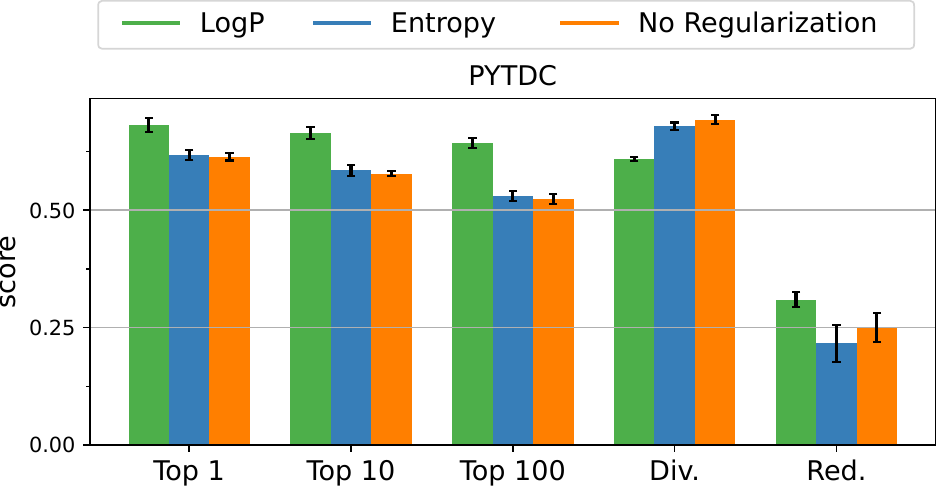}
    \vspace{-2em}\caption{\small  \label{fig:likelihood_smiles} Different likelihood penalization for exploration. Log P regularization is a better choice for efficient exploration for \methodName.}
    \vspace{-1em}
\end{wrapfigure}
\paragraph{Regularizing the policy's likelihood for exploration.} RL agents classically face an exploration-exploitation dilemma, which can lead to agents getting stuck in sub-optimal local maxima when not well balanced. \methodName agents are not immune to this dilemma. Upon encountering good, but sub-optimal molecules, an agent may adjust its policy to increase the likelihood of sampling these sub-optimal molecules and, without sufficient exploration, fail to discover higher-value regions of policy space. This can be particularly detrimental during the initial learning stages. 

To combat this issue, entropy regularisation, which adds a $\log\pi(s)$ term to the RL loss, has been proposed ~\citep{haarnoja2018soft}. This encourages the RL policy to explore states with lower likelihood values. Similarly \citep{olivecrona2017molecular} adds a Log p regularizer, which penalizes higher likelihood values by adding a $-1 / \log\pi(s)$ term to the RL loss. In~\Cref{fig:likelihood_smiles}, our results show that although an entropy regularizer leads to lesser redundancy, the Log p regularizer boosts performance significantly by exploring more efficiently. The Log p regularizer only penalizes the agent for being extremely certain (likelihood $\xrightarrow{\text{tends to}} 1$) about its actions, and is mostly agnostic for lower likelihood values. This penalty is a much better choice for \methodName as it only activates when stuck in a local optimum of molecular space. 

\section{Conclusion and Future Work}
We present \methodName that resulted from our empirical study of multiple algorithmic components of text-based molecular design. For future practitioners, our method suggests the following philosophy: (1) Using SMILES is a better choice than SELFIES. (2) When collecting data for pretraining, the quality of molecules matter much more than the number of molecules. (3) Both transformer and RNN architectures achieve similar performance across all tasks using current datasets. (4) Incorporating components such as a hill-climb buffer and Log P regularization yields substantial performance improvements. Conversely, introducing KL regularization or opting for more intricate actor-critic algorithms may result in diminished performance, at the cost of more hyperparameters and memory resources. While our analysis addresses many questions, it also shows that RL agents were able to \textit{hack} the reward functions suggesting that there is space to improve on the metrics used for molecule quality.

\bibliography{neurips_2023}
\bibliographystyle{plainnat}

\appendix
\clearpage

\paragraph{Outline of Appendices.} In~\autoref{appendix:experimental-setup} we provide details about the experimental setup. In~\autoref{appendix:hp-tuning} we describe our hyperparameter tuning strategy. In~\autoref{appendix:results} we include additional results from our experiments.

\section{Experimental setup}
\label{appendix:experimental-setup}
In this section, we provide additional details about the tasks, evaluation metrics and pretraining models and data used in our work.

\subsection{Tasks}
\label{appendix:tasks}
\paragraph{Pydtc tasks.} These tasks are a set of $21$ pharmaceutically-relevant oracle functions, which have been commonly used in prior work~\citep{brown2019gmole, gao2022pmo, huang2021therapeutics} for evaluating performance across molecular discovery algorithms:
\begin{itemize}
	\item QED: A quantitative estimate of drug-likeness calculated using a set of rules.
	\item DRD2, GSK3$\beta$, and JNK3: Classical machine learning models (SVMs and random forests) that provide an estimate of properties like target affinity or susceptibility towards a disorder.
	\item Celecoxib, Troglitazone, and Thiothixene rediscovery: An estimate of smiles text similarity, based on tanimoto metric, towards a target molecule.  
	\item Albuterol and Mestranol similarity: Generate molecules similar to a target molecule.
	\item Isomers\_c7h8n2o2 and isomers\_c9h10n2o2pf2cl: Generate molecules corresponding to a target molecular formula.
	\item Median1 and Median2: Generate molecules that are maximally similar to several target molecules.
	\item Osimertinib\_mpo, fexofenadine\_mpo, ranolazine\_mpo, perindopril\_mpo, amlodipine\_mpo, sitagliptin\_mpo, zaleplon\_mpo: Generate molecules that maximize multiple properties of a targeted drug.
	\item valsartan\_smarts: Generate molecules that contain a certain SMARTS pattern and certain physicochemical properties.  
\end{itemize}

Most of these tasks are from the GuacaMol benchmark~\citep{brown2019gmole}. All oracles are calculated using the Python API provided by Therapeutics Data Commons~~\citep{huang2021therapeutics} and more details for these tasks can be found on \href{https://tdcommons.ai/functions/oracles/}{their website}.

\paragraph{Docking tasks.} We used QuickVina 2~\citep{amr2015qvina} for calculating docking scores using the same default configuration parameters as prior works~\citep{yang2021hit, lee2023exploring}. For example, we used $\text{exhaustiveness} = 1$, and $ \text{modes} = 10$. We choose 5 different protein targets to calculate docking scores: fa7 (FA7), parp1 (PARP-1), 5ht1b (5-HT1B), jak2 (JAK-2), and braf (BRAF). These targets were chosen by~\citep{yang2021hit, lee2023exploring} because the docking simulators for these targets work fairly well when compared to the ground truth. In our experiments in ~\Cref{result:string-representations-architectures}, we found that text-based RL algorithms were easily able to produce chemically trivial molecules that have very high docking scores. To understand the complexity of computing docking scores, we report the time taken to dock 1000 molecules in parallel using 12 CPUs~\cref{table:docking-time}. We also provide the time taken to run the RL algorithm on these 1000 molecules after their docking scores are available.

\setlength{\tabcolsep}{4pt}
\begin{table}[ht]
\centering
\caption{\footnotesize \textbf{Time complexity of docking score evaluation} : More than half the running time is spent evaluating the docking scores.}
{\footnotesize
\begin{tabular}{l|c|c} \toprule
Number of molecules & Docking time & RL update time\\
\midrule
1000 & 130 seconds
& 74 seconds \\
\bottomrule
\end{tabular}
}
\label{table:docking-time}
\end{table}

\paragraph{Augmented docking tasks.} In our results for the standard \textbf{docking} tasks~(\autoref{fig:smiles-vs-selfies} and~\autoref{fig:architecture_smiles}), we found that using the simulated docking scores as rewards did not lead to chemically relevant molecules. Text-based RL algorithms were able to exploit their state and action spaces to design chemically trivial molecules that have very high docking scores. To tackle this issue of undesirable reward hacking, we tried a reward function based on prior works~\citep{garcia2022dockstring, lee2023exploring} that combine objectives for drug-like, and synthesizable molecules with docking scores. We call tasks corresponding to this new reward function as \textbf{augmented docking} tasks. Concretely, we chose the same reward function from~\citep{lee2023exploring}

\begin{equation}
	r(s) = -\text{DS}(s)/20 \times \text{QED}(s) \times (10 - \text{SA}(s))/9,
\end{equation}

Where DS is the docking score, QED and SA are quantitative estimates of drug likeness and synthesizablity respectively.

\subsection{Evaluation metrics}
\label{appendix:evaluation-metrics}
Most of the metrics we use are described in detail in ~\Cref{sec:experiment-setup}. Here, we provide additional details about the diversity metric. We calculate the diversity of the top $100$ molecules sampled by the algorithm, where higher diversity is considered better given that it increases the chances for success in further wet lab experiments. In our experiments, we use the diversity evaluator from TDC~\citep{huang2021therapeutics}, which defines the diversity of a set of molecules as the average pairwise Tanimoto similarity between Morgan fingerprints of the molecules. See Section 2 of~\citep{benhenda2017chemgan} for exact details of how Tanimoto similarity is calculated.

\subsection{Pretraining}
\label{appendix:pretraining}
In this section, we provide more details about the pretraining datasets and models used in our experiments.

\paragraph{Pretraining datasets.} The ZINC $250$k dataset contains approximately $250$k molecules from the ZINC database \citep{irwin2012zinc}, chosen for their pharmaceutical relevance, moderate size, and popularity~\citep{gao2022pmo}. The CHEMBL dataset~\citep{mendez2019chembl} consists of approximately 2M manually curated drug-like molecules. The other $3$ datasets consist of randomly selected subsets of the ZINC-$15$ dataset \citep{sterling2015zinc} that obey some chemically imposed mild constraints \citep{irwin2022chemformer}. We test three subsets of different sizes: (1) ZINC $1$M  (2) ZINC $10$M, and (3) ZINC $100$M, to test the impact of scaling the size of pre-training data. These datasets and data-subsets, including their vocabularies, will be shared in an easily accessible format upon acceptance.

Removing outliers and unusual non drug-like compounds helps to keep the vocabulary small and improves the quality of the generative model~\citep{blaschke2020reinvent}. To achieve this, we filter all datasets by removing molecules which contain 1) less than $10$ or more than $50$ heavy atoms and 2) molecules \textit{other than} Carbon, Nitrogen, Oxygen, Fluorine, Silicon, Chlorine and Bromine. 
We also canonicalize and sanitize all molecules using RDKIT \citep{landrum2013rdkit}. For experiments that apply SELFIES, we convert all datasets to SELFIES using the Python API provided by~\citep{krenn2020selfies} (Version: 2.1.1). 

Apart from the experiments shown in the main paper, \Cref{appendix:pretraining-exp} contains additional experiments comparing text-based RL agents across different pretraining datasets.

\paragraph{Pretraining models.} In~\autoref{table:pretraining-model} we provide details about the pretraining modes which we use in our experiments. Upon acceptance, we will open-source our code and release the pretrained weights to support reproducible research. 

\setlength{\tabcolsep}{4pt}
\begin{table}[ht]
\centering
\caption{\footnotesize \textbf{Description of model architectures used for pretraining}}
{\footnotesize
\begin{tabular}{l|c|c} \toprule
Model & Number of Parameters & Description\\
\midrule
FC & \num{10674224} 
& \makecell{ FC is a fully connected neural network \\ with 3 hidden layers of $1024$ size each.} \\
\midrule
RNN & \num{4170034} & \makecell{RNN is a recurrent network which consists of \\ 3 GRU layers of hidden sizes $512$ each.} \\
\midrule
TRANSFORMER & \num{4782384} & \makecell{GPT~\citep{brown2020language} style transformer with 6 layers, \\ 16 heads and $256$ embedding dimensions.} \\
\bottomrule
\end{tabular}
}
\label{table:pretraining-model}
\end{table}

We select network sizes that have been commonly used in RL~\citep{blaschke2020reinvent, yarats2021drqv2}. Although conducting a study of scaling the model size~\citep{kaplan2020scaling} is out of the scope of our work, we believe that it is a promising direction for future.

Since the fully connected model can only take fixed length inputs, we always input a molecular text padded to a certain maximum length (we used length 100 in our experiments). This padding is done using a special token [PAD] to convey that corresponding tokens should not be considered while deciding the value of the text.

\paragraph{Pretraining experimental details.} We pretrain FC, RNN and transformer architectures on the ZINC $250$K dataset and pretrain a transformer on all other datasets. All models are pretrained using the PyTorch~\citep{paszke2019pytorch} framework. All models used an initial learning rate of $1e-3$, with a cosine learning rate schedule~\citep{loshchilov2017sgdr}. FC and RNNs used a batch size of 128 and were trained for 10 epochs. All transformers were trained for 5 epochs, with the largest batch size that we could fit in the memory of a single NVIDIA RTX A$6000$ GPU, for example, a batch size of $2048$ for pretraining the transformer on ZINC $100$M dataset. We made sure that all models were trained until convergence. On the ZINC $250$K SMILES dataset, the FC, the RNN and the transformer model achieved a validation loss of 29.417, 22.507, and 22.923 respectively. 

\subsection{RL finetuning}
\label{appendix:rl-finetuning}
The pretrained model is further trained using the policy gradient algorithm, REINFORCE~\citep{sutton1999reinforce}. Given the reward function $r(s_H)$ corresponding to the text $s_T$, this algorithm optimizes the loss function

\begin{equation}
 \min_{\theta} J(\theta) = - \left[ \sum_{t=1}^{H} \log p_{\theta}(a_t=A_t \mid A_{t-1}, \cdots A_0)  r(s_H = [A_{0}, \cdots A_H] ) \right],
\end{equation}

where $A_t$ is the token sampled by the agent at time-step $t$.

\section{Hyperparameter tuning}
\label{appendix:hp-tuning}
We conduct a common hyperparameter tuning strategy for all experiments. Specifically, we conduct hyperparameter tuning for
\begin{itemize}
	\item Learning rate for different architectures~\autoref{fig:architecture_smiles} and text grammars~\autoref{fig:smiles-vs-selfies}.
	\item Coefficients for different likelihood regularizations~\autoref{fig:likelihood_smiles}.
	\item Coefficients for KL regularization loss term~\autoref{fig:prior-kl}.
\end{itemize}

We select three tasks from the \textbf{pytdc} tasks, i.e., troglitazone\_rediscovery, sitagliptin\_mpo, and median2 for hyperparameter tuning. For each hyperparameter, we select a set of 5 evenly spaced realistic values and run 5 random seeds of RL experiments per hyperparameter value. We select the hyperparameter value that achieves the best average score of the top-$100$ molecules as the final value for running all the experiments. We report the hyperparamters used for the policy gradient training in~\cref{table:hyperparamters}.

\setlength{\tabcolsep}{4pt}
\begin{table}[ht]
\centering
\caption{\footnotesize \textbf{Hyperparamters}}
{\footnotesize
\begin{tabular}{l|c} \toprule
Name & Value\\
\midrule
Maximum number of unique molecules & 25000 \\
\midrule
Learning rate & \makecell{\num{0.0005} RNN and FC \\ \num{0.0001} Transformer}  \\
\midrule
Batch size & 64  \\
\midrule
Log p coefficient & 5  \\
\midrule
KL coefficient & \num{0.001} \\
\bottomrule
\end{tabular}
}
\label{table:hyperparamters}
\end{table}

\section{Results}
\label{appendix:results}

\subsection{Text representations and architectures for RL}
\label{appendix:string-representations-architectures}
\begin{figure}[ht]
    \begin{subfigure}[b]{1\textwidth}
    \centering
    \includegraphics[width=1\textwidth]{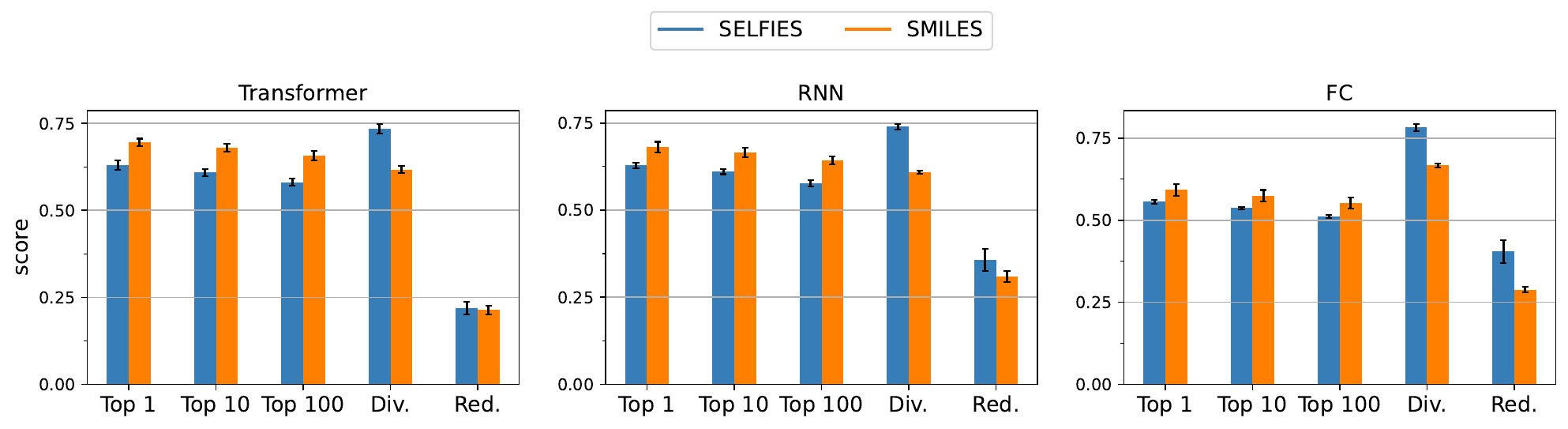}
    \subcaption{PYTDC.}
    \end{subfigure}
    
    \begin{subfigure}[b]{1\textwidth}
    \centering
    \includegraphics[width=1\textwidth]{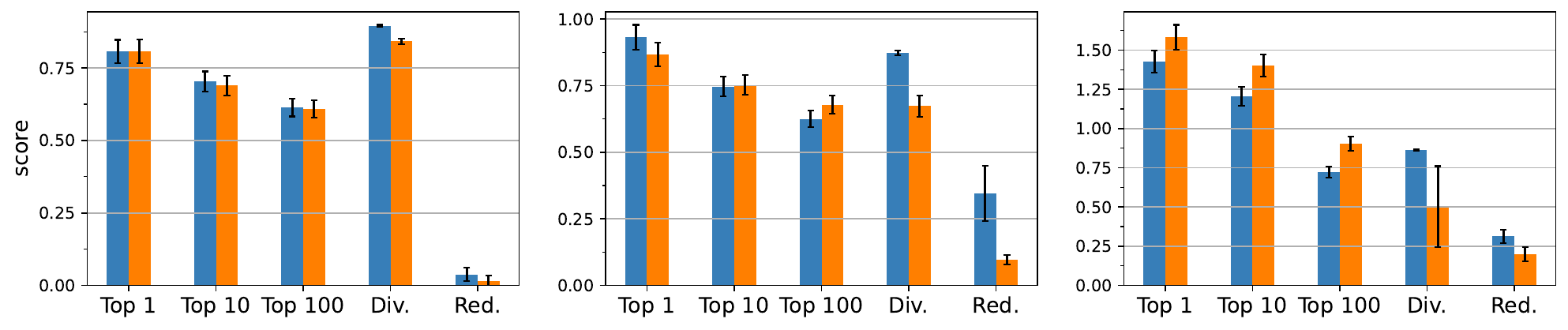}
    \subcaption{DOCKING.}
    \end{subfigure}
    \caption{\label{fig:appendix-smiles-vs-selfies} Comparison between SELFIES and SMILES across different architectures. These figures are the individual plots corresponding to the normalised plot show in~\autoref{fig:smiles-vs-selfies}.}
\end{figure}

\begin{figure}[ht]
    \centering
    \includegraphics[width=1\textwidth]{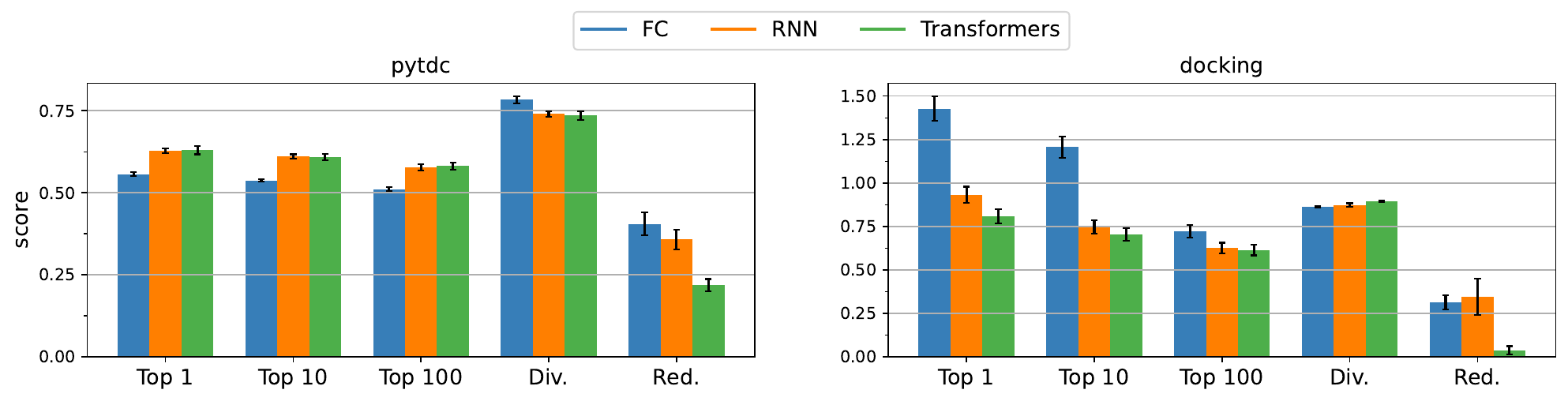}
    \caption{\label{fig:appendix-architecture_selfies} \textbf{Comparison of different policy architectures (SELFIES):} No single architecture clearly outperforms for molecular text-based RL. Although FC does better on the docking tasks, our analysis shows that it learns to exploit the docking function as opposed to designing high-value molecules.}
\end{figure}

Here, we present additional results from~\autoref{result:string-representations-architectures}. \autoref{fig:appendix-smiles-vs-selfies} shows that SMILES are a better molecular grammar when compared to SELFIES across all architectures, for the text based RL algorithms that we consider. \autoref{fig:appendix-architecture_selfies} compares various architectures, while keeping the molecular grammar fixed to SELFIES. The results in~\autoref{fig:appendix-architecture_selfies} reflect our findings in~\autoref{fig:architecture_smiles} that no single architecture clearly outperforms for molecular text-based RL. It also shows the reward hacking behavior of the \textbf{docking} tasks by the FC based RL agent.

The reason for lower value molecules for SELFIES environments can be explained by the SELFIES grammar that induces a flat optimization landscape. Many SELFIES strings can correspond to the same molecule, and in fact, once an \textit{invalid} action is taken, any subsequent sequence of tokens will be ignored on the resulting molecule. This makes exploration of new molecules difficult~\citep{krenn2020selfies, gao2022pmo}. On the other hand, the benefit of SELFIES over SMILES in eliminating invalid molecule generation is mitigated by our pretraining process, which initializes SMILES-based policies with a strong bias toward generating valid molecules. Overall, we find that SMILES-based policies, when combined with pretraining, are more effective at exploring and finding high-value molecules.  

\subsection{Pretraining for RL}
\label{appendix:pretraining-exp}
\begin{figure}[ht]
    \centering
    \includegraphics[width=1\textwidth]{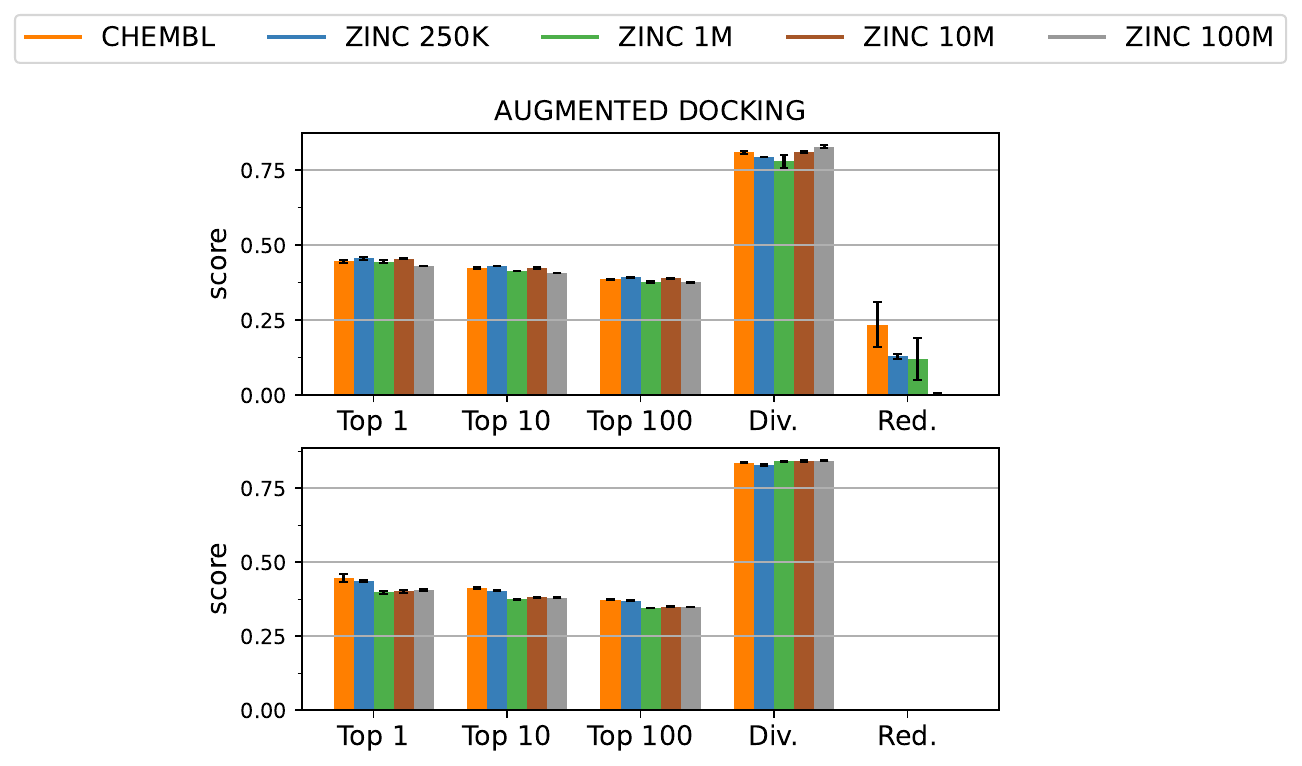}
    \caption{\label{fig:appendix-prior-data} This figure shows the augmented rewards obtained by the RL agents (Top) and data quality (Bottom) of different datasets. See~\autoref{appendix:tasks} for how the augmented reward is calculated.}
\end{figure}

\autoref{fig:prior-data} (right) shows the top docking scores obtained by RL agents pre-trained on different datasets when trained with on the \textbf{augmented docking} tasks. In~\autoref{fig:appendix-prior-data}, we show the actual augmented rewards obtained by the RL agent. These results suggest that the augmented docking score is a complex reward function as the RL agent is achieved minimal improvement over the prior agent. To verify this hypothesis, we increased the molecule budget of the RL agent by 10 times. We indeed see that RL agents corresponding to all prior-datasets exhibit considerable improvement. Text-based RL algorithms learn to search more efficiently when provided with more compute.

\begin{figure}[]
    \centering
    \includegraphics[width=1\textwidth]{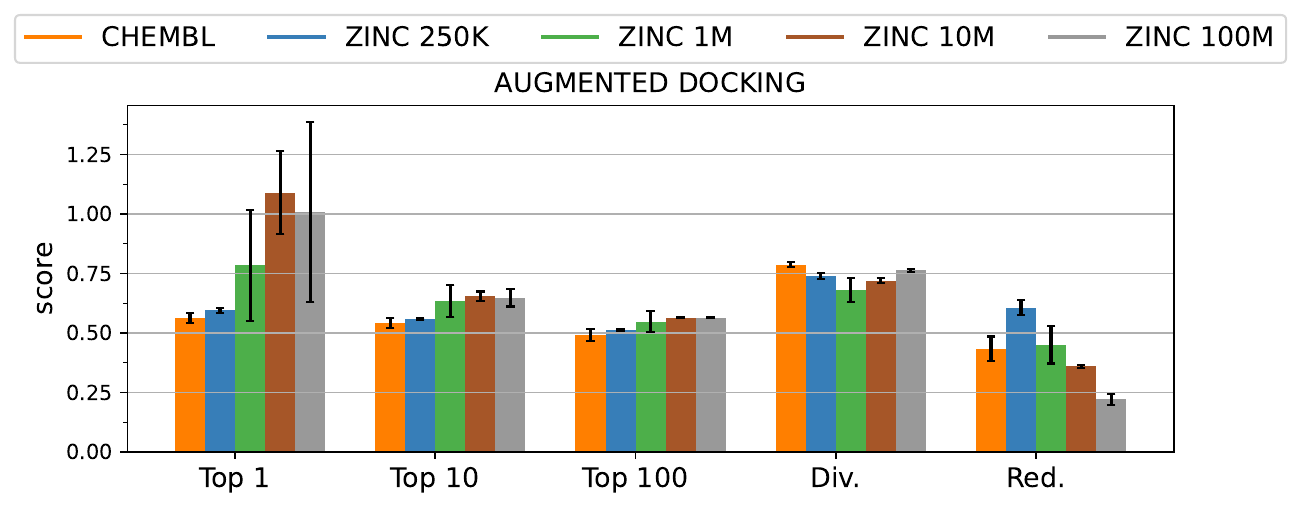}
    \caption{\label{fig:appendix-10x-prior-data} This figure shows the augmented rewards obtained by the RL agents trained for 10 times more molecules.}
\end{figure}

\subsection{Reward Hacking}
\label{appendix:reward-hacking}

\autoref{fig:architecture_smiles} and~\autoref{fig:appendix-architecture_selfies} show that text-based RL agents that are trained using fully connected neural networks are able to obtain unusually high rewards. This is probably because it is easier for FC agents to find actions that exploit the local structure of the reward function as RNNs and Transformers are inductively biased to find global solutions. This highlights an undesirable type of reward function hacking by the FC agent which provides high rewards for molecules with long strings of Carbon and Nitrogen atoms together. Similar to prior work~\citep{lee2023exploring}, we augment the docking scores with objectives for drug-like and synthesizable molecules. See~\autoref{appendix:experimental-setup} for details of this task and~\autoref{fig:prior-data} and~\autoref{fig:appendix-prior-data} for results corresponding to this task. Our initial results on this task (~\autoref{fig:prior-data} and~\autoref{fig:appendix-prior-data}) suggested that the augmented reward function was more aligned towards chemically relevant molecules. We also noticed that the RL agents were not able to improve a lot over the prior baselines for this task. To verify whether the low performance of RL agents was because of less training data or the augmented reward function was indeed a more realistic and robust reward function, we repeated the experiments in~\autoref{fig:appendix-prior-data} with a ten times higher training budget. Given more data, all RL agents showed considerable improvements over the priors. This experiment also revealed that the agents pre-trained on ZINC $1$M, ZINC $10$M and ZINC $100$M, were able to exploit the reward function to generate unrealistic yet highly rewarding molecules. These molecules have unusually low docking scores (less than -20). Our results highlight the need for an aligned and a more robust reward function to generate molecules for docking protein targets. 

\subsection{Additional results for the importance of algorithmic choices for text-based RL.}
\label{appendix:algo-comp}

\begin{figure}[ht]
    \centering
    \includegraphics[width=0.9\textwidth]{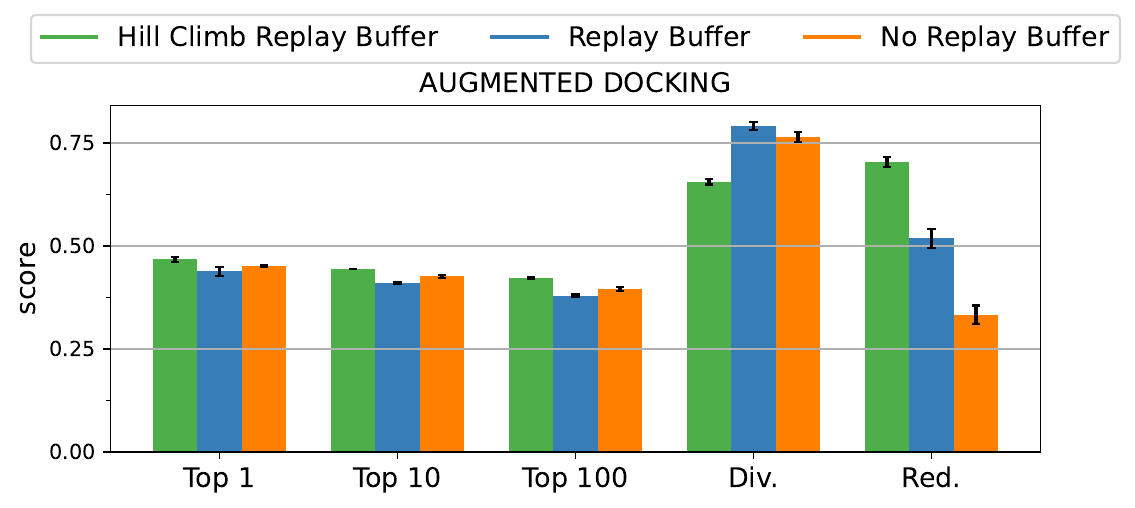}
    \caption{\label{fig:appendix_replay_buffer_smiles} \textbf{Do replay buffers help? }}
\end{figure}

\begin{figure}[ht]
    \centering
    \includegraphics[width=0.9\textwidth]{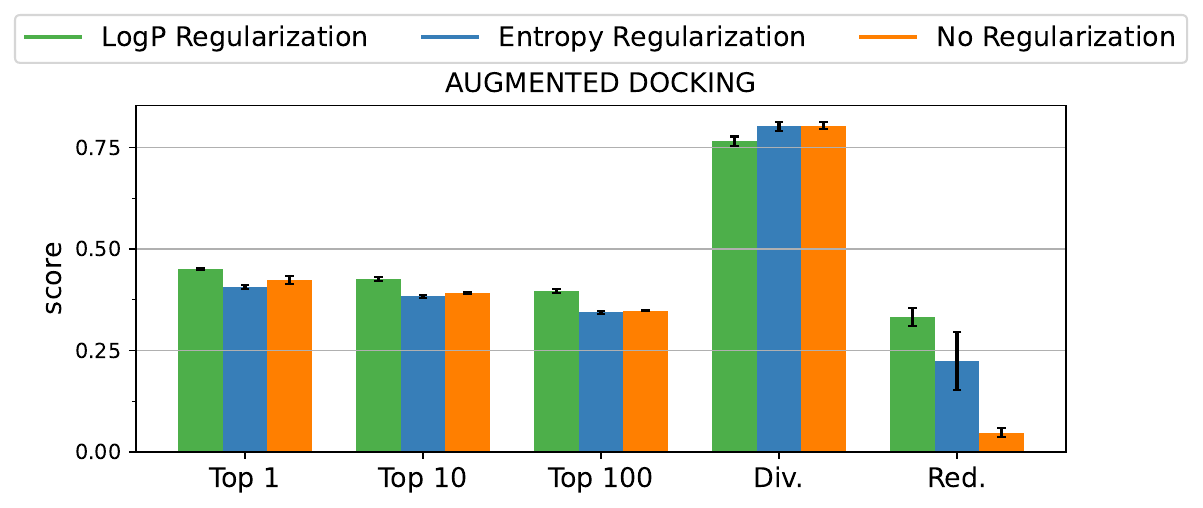}
    \caption{\label{fig:appendix_likelihood_smiles} \textbf{Comparison of different likelihood penalization for efficient exploration}}
\end{figure}

\begin{figure}[ht]
    \centering
    \includegraphics[width=0.9\textwidth]{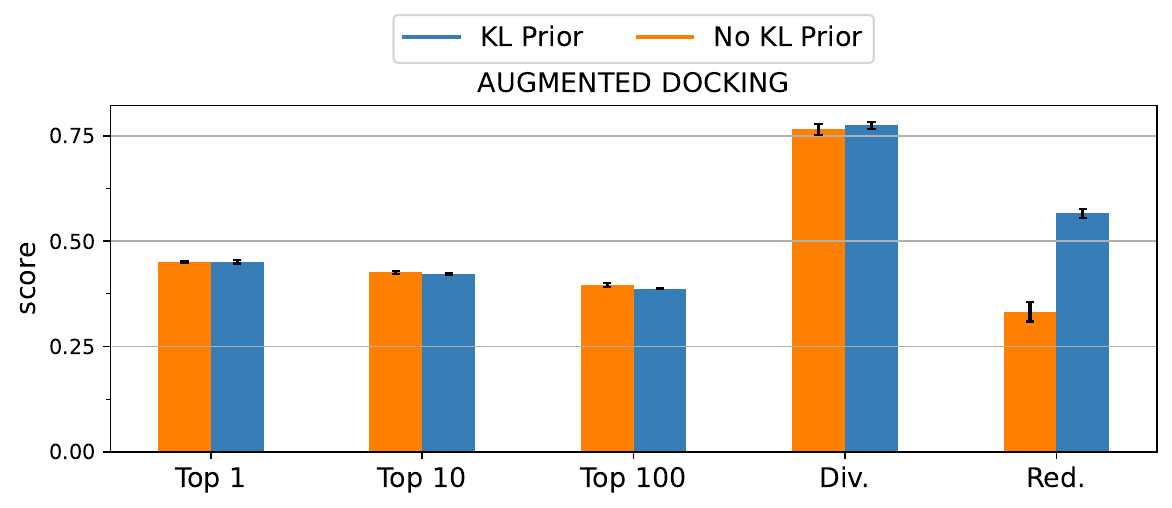}
    \caption{\label{fig:appendix_kl_prior_smiles} \textbf{Is KL regularisation with a prior necessary?}}
\end{figure}

\Cref{result:algorithm-tricks} compares various algorithmic components like replay buffers, hill climbing, KL regularisation towards the pretrained policy, and likelihood penalties and show results for \textbf{PYTDC} tasks. In this section, we repeat all the experiments from~\Cref{result:algorithm-tricks} on \textbf{augmented docking} tasks as well and reach the same conclusions. In~\autoref{fig:appendix_replay_buffer_smiles} we see that using the hill-climb buffer results in a significant performance boost, whereas using a standard buffer does not contribute much.~\autoref{fig:appendix_likelihood_smiles} shows that Log P regularization is a better choice for efficient exploration when it comes to text-based RL algorithms. In~\autoref{fig:appendix_kl_prior_smiles} show that penalising the policy to move away from the pretrained policy does not improve performance.

\subsection{Instability of transformers for online RL.}
\label{appendix:transformer-instability-rl}
\begin{figure}[]
    \centering
    \includegraphics[width=1\textwidth]{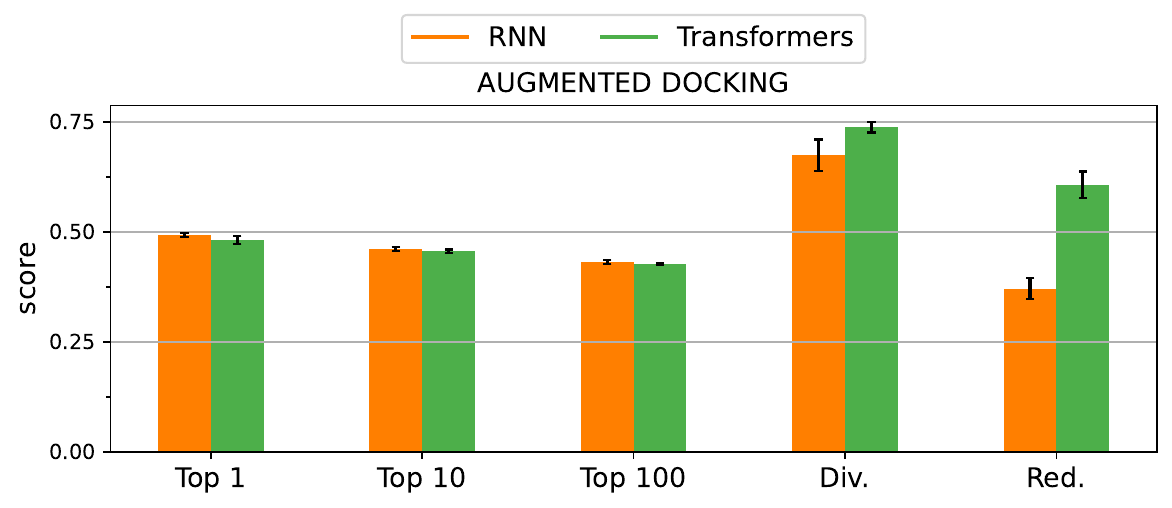}
    \caption{\label{fig:appendix-10x-architecture} A transformer and an RNN based RL agent trained for 10 times more molecules on augmented docking scores.}
\end{figure}

Many works~\citep{parisotto2019stabilizing} have pointed out the instability of training transformers using online reinforcement learning. To understand this in the context of text based RL, we compare a transformer and an RNN based agent on the augmented docking task. To probe whether pronounced effects of this instability are seen, we train both agents for 10 times more molecules ($250$K molecules). In figure~\autoref{fig:appendix-10x-architecture}, we see that both agents perform comparably across all docking targets. 

\begin{figure}[]
    \centering
    \includegraphics[width=1\textwidth]{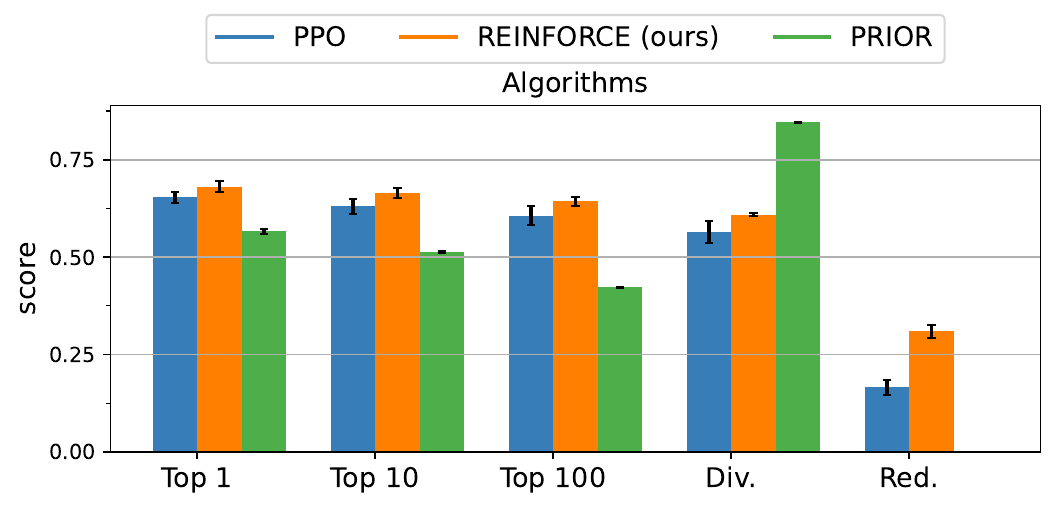}
    \caption{\label{fig:ppo-pytdc} \textbf{Comparison with another RL algorithm PPO.} In control tasks complex algorithms like PPO~\citep{schulman2017proximal} are known to outperform the vanilla policy gradient algorithm. But on the molecular optimization tasks of PyTDC, our results indicate that vanilla policy gradient algorithms are more stable than actor critic algorithms like PPO and achieve higher performance. This resonates with the findings of previous work in molecular optimization~\citep{cieplinski2021able, gao2022pmo}.}
\end{figure}

\end{document}